\documentclass{article}

\usepackage[final]{corl_2018} 
\usepackage{epsfig}
\usepackage{amsthm}
\usepackage{lineno,hyperref}
\modulolinenumbers[5]
\usepackage{graphicx, wrapfig}
\usepackage{amssymb}
\usepackage{amsmath}
\usepackage{algorithm}
\usepackage[noend]{algpseudocode}
\usepackage{mathptmx} 
\usepackage{amsmath}
\usepackage{subfig}
\usepackage{float}
\usepackage{mathtools}
\usepackage{booktabs}

\makeatletter

\newcommand{\Rmnum}[1]{\expandafter\@slowromancap\romannumeral #1@}
\def\BState{\State\hskip-\ALG@thistlm}
\makeatother

\title{Attention based visual analysis for fast grasp planning with multi-fingered robotic hand}

\author{
  Zhen Deng\\
  University of Hamburg 
  United States\\
  \texttt{deng@informatik.uni-hamburg.de} \\
  \And
  Ge Gao \\
  University of Hamburg 
  United States\\
  \texttt{gao@informatik.uni-hamburg.de} \\
	\And
  Simone Frintrop \\
  University of Hamburg 
  United States\\
  \texttt{frintrop@informatik.uni-hamburg.de} \\
	\And
  Jianwei Zhang \\
  University of Hamburg 
  United States\\
  \texttt{zhang@informatik.uni-hamburg.de} \\
}
\begin{document}
\maketitle


\begin{abstract}
We present an attention based visual analysis framework to compute grasp-relevant information in order to guide grasp planning using a multi-fingered robotic hand.
Our approach uses a computational visual attention model to locate regions of interest in a scene, and uses a deep convolutional neural network to detect grasp type and point for a sub-region of the object presented in a region of interest. We demonstrate the proposed framework in object grasping tasks, in which the information generated from the proposed framework is used as prior information to guide the grasp planning.
Results show that the proposed framework can not only speed up grasp planning with more stable configurations, but also is able to handle unknown objects.
Furthermore, our framework can handle cluttered scenarios.
A new Grasp Type Dataset (GTD) that considers 6 commonly used grasp types and covers 12 household objects is also presented.
 \end{abstract}


\section{Introduction}
\label{sec:intro}
Imagine a toddler is in front of a tabletop with several objects, very likely he or she would interact with those objects by trying to pick up the red mug either by the handle or the rim, or trying to grasp the green ball. 
The ability of rapidly extracting relevant information from visual input is an important mechanism and a natural behavior for humans to conduct various activities.
The majority of visual analysis approaches for grasp planning with multi-fingered robotic hands follow a pipeline containing object localization, object representation and recognition \cite{schwarz2017nimbro}.
However, reliable object detectors such as deep-learning based approaches require huge amounts of training data, as well as good hardware to achieve a reasonable time performance for robotic applications, while handcrafted feature based approaches can not handle the dynamics in real life scenarios. 

This paper proposes an attention based visual analysis framework which directly locates sub-regions of objects as regions of interest (ROIs), and generates grasp-relevant information from visual data inside ROIs for grasp planning by a multi-fingered robotic hand.
The proposed learning framework is inspired by psychological studies which demonstrated that humans combine an early bottom-up processing with a later top-down processing to visually analyze the scene \citep{theeuwes2010top, awh2012top}. 
The bottom-up process starts with sensory input data and is completely stimulus-driven, while the top-down process extracts relevant information, which may be influenced by prior experience and semantics.
In particular, a computational attention model is used to process visual data and outputs a pixel-precise saliency map, from which salient regions are selected for further processing. 
Meanwhile, the grasp type and point on the object segments presented in salient regions is predicted by a network.  
Finally, this information is used to guide the grasp planning with a multi-fingered robotic hand. 


Grasp type and point convey useful information for planning the configuration of a robotic hand. 
However, many previous works on grasp type detection mainly focus on the analysis of human hand behavior \citep{rogez2015understanding, cai2017ego}, while few approaches integrate grasp type detection into robotic grasp planning. 
Furthermore, those works that consider grasp types for robotic hands normally divide grasp types into power and precision \citep{napier1956prehensile} and decide the grasp type manually during planning, which is not sufficient for exploring the potential of multi-fingered robot hands.
In terms of visual analysis, there are approaches in \citep{hsiao2010contact,aleotti20123d,vahrenkamp2018planning}, which use visual analysis to define heuristics or constraints for grasp planning. 
In comparison to those approaches, there are two main differences: 1) our approach learns features directly from raw sensory data, while most of the previous approaches use handcrafted features; 2) 6 grasp types are considered while the previous approaches only consider 2 grasp types. 
To the best of our knowledge, this is the first approach which integrates grasp type detection into grasp planning for multi-fingered robotic hands.

In this paper, we address the problem of visual analysis of natural scenes for grasping by multi-fingered robotic hands. 
The objective is to compute grasp-relevant information from visual data, which is used to guide grasp planning. 
A visual analysis framework which combines a computational visual attention model and a grasp type detection model is proposed. 
A new Grasp Type Dataset (GTD) which considers six commonly used grasp types and contains 12 household objects is also presented.

The rest of the paper is organized as follows. 
Section \ref{sec:related_work} presents related work. 
Section \ref{sec:system_architecture} introduces the architecture and main components of the proposed visual analysis framework. 
Grasp planning is described in Section \ref{sec: grasp_planning}. 
Experimental results are presented in Section \ref{sec:result}. 
Finally, the conclusion and future work are discussed in Section \ref{sec:conclusion}.
\section{Related Work}
\label{sec:related_work} 
Information extracted from visual analysis can be used to define heuristics or constraints for grasp planning.
Previous grasp planning methods can be divided into geometric-based grasping and similarity-based grasping. 
In geometric-based grasping \citep{hsiao2010contact,laga2013geometry,vahrenkamp2018planning}, geometric information of the object is obtained from color or depth images, and is used to define a set of heuristics to guide grasp planning.
\citet{hsiao2010contact} proposed a heuristic which maps partial shape information of objects to grasp configuration. 
The direct mapping from object geometric to candidate grasps is also used in \citep{harada2008fast, vahrenkamp2018planning}. 
\citet{aleotti20123d} proposed a 3D shape segmentation algorithm which firstly oversegments the target object, and candidate grasps is chosen based on the shape of the resulting segments \citep{laga2013geometry}. 
In similarity-based approaches \citep{herzog2014learning,dang2014semantic,kopicki2016one}, the similarity measure is calculated between the target object and corresponding object model from human demonstrations or simulation. 
The candidate grasp is then queried from datasets based on similarity measures. 
\citet{herzog2014learning} defined an object shape template as the similarity measure. 
This template encodes heightmaps of the object observed from various viewpoints. 
The object properties can also be presented with semantic affordance maps \citep{dang2014semantic} or probability models \citep{kopicki2016one, kroemer2014predicting}. 
Geometric-based approaches usually require a multiple-stage pipeline to gather handcrafted features through visual data analysis. 
Due to vision sensory noise, the performance of the geometric-based grasping is often unstable. 
Meanwhile, similarity-based methods are limited to known objects which can not handle unknown objects. 
In contrast to previous methods, our method increases grasping stability by learning more reliable features, meanwhile it is able to handle unknown objects.

Many saliency approaches have been proposed in the last two decades. 
Traditional models are usually based on the feature integration theory (FIT) \citep{Treisman1980FIT} to compute several handcrafted features which were fused to a saliency map, e.g the iNVT  \citep{itti1998,Dirk2006iNVT} or the VOCUS system \citep{Frintrop2006vocus}.
\citet{frintrop2015} proposed a simple and efficient system which computes multi-scale feature maps using Difference-of-Gaussian (DoG) filters for center-surround contrast and produces a pixel precise saliency map. 
Deep learning based saliency detection mostly relies on high-level pretrained features for object detection tasks due to the requirements of massive amounts of training data \citep{huang2015salicon, Lietal2016deepsaliency, liu2016DHSNet}. 
\citet{Kümmerer2014b} used an AlexNet \citep{Krizhevsky2012Alexnet} pretrained on Imagenet \citep{Deng2009ImageNetAL} for object recognition tasks, the resulting high-dimensional feature space is used for fixation prediction and saliency map generation. 
Since most of the deep-learning based approaches have a central photographer bias which is not desired in robotic applications, we choose to use a handcrafted feature based approach which gathers local visual attributes by combing low-level visual features \citep{frintrop2015}.


\section{Attention based visual analysis}
\label{sec:system_architecture}
\begin{figure}
\centering
\includegraphics[width=0.8\textwidth]{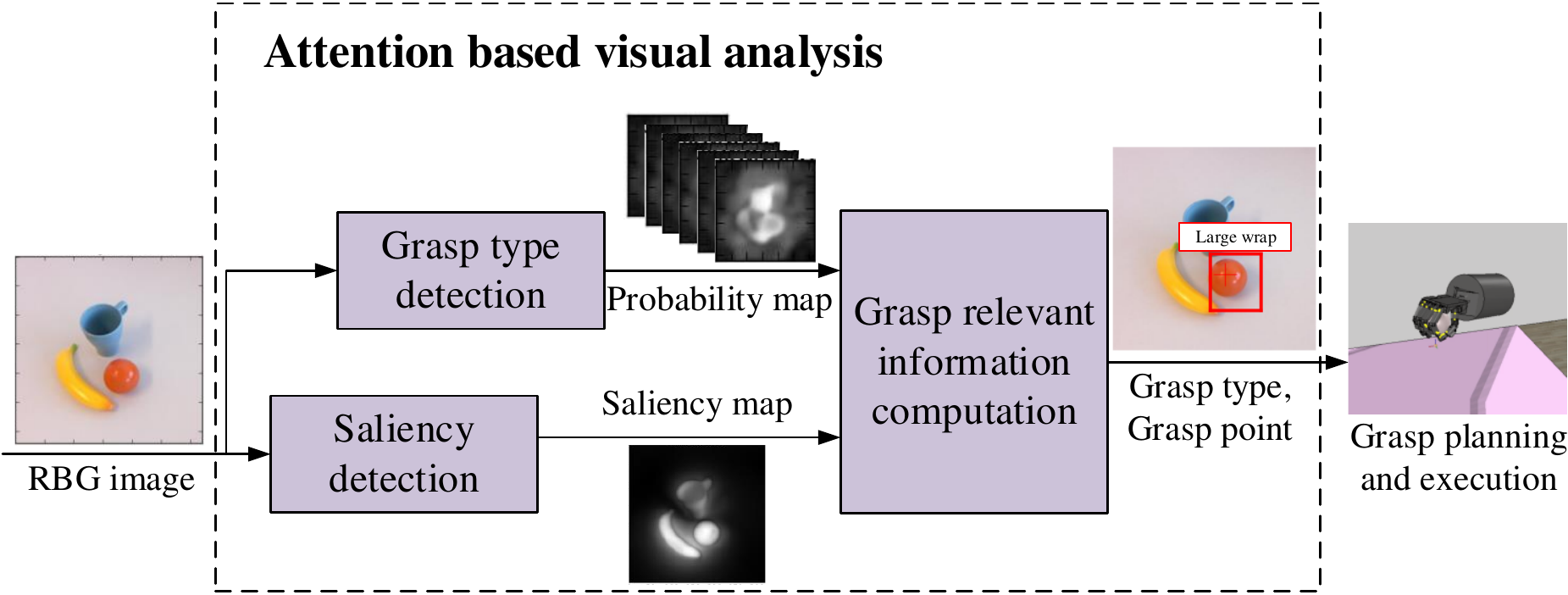}
\caption{The proposed attention based visual analysis framework. 
With an input RGB image, a ROI is selected using the saliency map produced by a \textit{Saliency detection} model. 
Inside the ROI, grasp type and point are computed based on the six probability maps produced by the \textit{Grasp type detection} network. 
The obtained information containing grasp type and point is then used as a prior to guide grasp planning.
The planned grasp is executed in a physical simulator to verify its quality.}
\label{fig:system_overview}
\end{figure}

The proposed framework contains two main components, a computational visual attention model which gathers low-level visual features and selects ROIs for further processing, and a grasp type detection model which learns higher level features and produces grasp-relevant information in the ROIs.
Figure \ref{fig:system_overview} illustrates an overview of the proposed attention based visual analysis framework. 

\subsection{Computational visual attention model}
\label{subsec:saliency_model}
The pixel-level saliency map is computed using the computational visual saliency method VOCUS2 \citep{frintrop2015}. 
In principle, any saliency system which has real-time capability and does not have a center-bias could be used. 
Center bias gives a preference to the center of an image, which is not desired in robotics applications. 
Unfortunately, this excludes most deep-learning based approaches since they are usually trained on large datasets of Internet images, which mostly have a central photographer bias. 
Therefore, the VOCUS2 system was chosen, which belongs to the traditional saliency systems with good performance on several benchmarks.
In VOCUS2, an RGB input image is converted into an opponent-color space including intensity, red-green and blue-yellow color channels. 
DoG contrasts are computed with twin pyramids, which consist of two Gaussian pyramids - one for the center and one for the surround of a region - which are subtracted to obtain the DoG contrast. 
Finally, the contrast maps are fused across multiple scales using arithmetic mean to produce the saliency map.
Salient regions that have a high contrast to the surroundings are clustered and selected and passed to the next stage for further processing \citep{theeuwes2010top}. 

\subsection{Grasp type detection}
\label{subsec:grasp_type_detection}
\begin{figure}[H]
\centering
\includegraphics[width=0.85\textwidth]{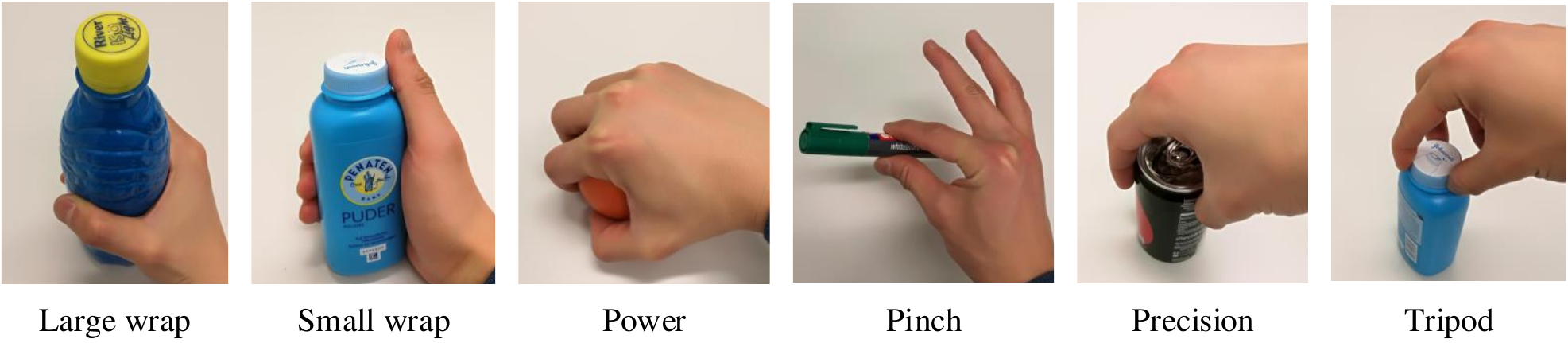}
\caption{The proposed 6 commonly used grasp types.}
\label{fig:grasp_type}
\end{figure}

Grasp type is a way of representing how a hand handles objects. 
Typically, the robotic grasps are divided into power and precision grasp \citep{napier1956prehensile}. 
Power grasp uses the fingers and palm to hold the object firmly, while precision grasp only uses fingertips to stabilize the object. 
However, this two-categories grasp taxonomy is not sufficient to convey information about hand configuration. 
\citet{feix2016grasp} introduced a GRASP taxonomy in which $33$ different grasp types used by humans are presented. 
Considering the kinematic limitations of the robotic hand as well as Feix's GRASP taxonomy, we extend the above two-categories grasp taxonomy into 6 commonly used grasp types: \textit{large wrap}, \textit{small wrap}, \textit{power}, \textit{pinch}, \textit{precision} and \textit{tripod}. 
Figure \ref{fig:grasp_type} illustrates the proposed grasp taxonomy. 
In order to detect grasp types directly from visual data, we refer to the architecture proposed by \citet{chen2018deeplab}. 
Since an object may have multiple feasible grasp types, the grasp type detection is a multi-label detection. Hence, we modify the output layer of the network and do not use the additional fully connected Conditional Random Field (CRF). 
Corresponding to 6 grasp types, the modified network predicts 6 pixel-level probability maps with the same resolution as the input image. 
In order to train the modified network for grasp type detection, this paper introduces a grasp type detection (GTD) dataset, in which 12 household objects are used and all the instances are annotated following the proposed 6 grasp types. 
The details of the GTD dataset and the model training are provided in Section \ref{subsec:dataset_and_implementation}.

Given an RGB image $I$ with height and width $h \times w$ as input, our network outputs pixel-level probability maps $P(Y|I)$ for each grasp type $s \in [1,2,\cdots, 6]$, where $Y = \{y_{i,j}^{s} \}_{i=1:h, j=1:w}$ . 
And the predicted probability of pixel $\{ [i,j]_{i=1:h,j=1:w}\}$ belonging to the grasp type $s$ is denoted by $y_{i,j}^{s}$. 
With the pixel-level probability maps, the best grasp type for the ROI is selected by summing the predicted probabilities of all the pixels inside ROI to obtain the probability $P(Y^{s}|O)$ of grasp type $s$ for ROI $O$, as shown in Eq. \ref{eq:grasp-region}. 
The grasp type  $s^{*}$ with the highest probabilities $P(Y^{s}|O)$ is chosen, in which $s \in [1,2,\cdots, 6]$: 
\begin{equation}
P(Y^{s}|O) = \frac{1}{h_{O} \times w_{O}} \sum_{i=1}^{h_{O}}\sum_{j=1}^{w_{O}} P( y_{i,j}^{s}| x_{i,j}),  \forall s \in [1,2,\cdots, 6].
\label{eq:grasp-region}
\end{equation}

After determining the best grasp type $s^{*}$, we need to localize the grasp point for the grasp type $s^{*}$ inside $O$. 
In order to find a stable grasp point $p$, subregions with higher predicted probabilities are clustered. 
Mean Shift \citep{comaniciu2002mean} is used to find a grasp point $p$ in $O$.
Multiple clusters with multiple centers are produced, and the cluster center with highest probability is selected as the grasp point $p$. 
Finally, the grasp relevant information $\Omega = \{O, s^{*}, p_{o}\}$, i.e., ROI $O$, the grasp type $s^{*}$ and point $p_{o}$, are generated from the proposed visual analysis framework.

\section{Grasp planning with prior information}
\label{sec: grasp_planning}
Information $\Omega $ generated from the proposed visual analysis framework is used as a prior for guiding the grasp planning. 
The objective of the grasp planning is to find a feasible grasp configuration for stable grasping. 
Using a multi-fingered robotic hand to grasp objects requires the computing of: a) a joint configuration of the robotic hand, b) contact points on the object surface and c) relative pose between the object and the robotic hand. 
Due to the high dimensionality of the robotic hand, it is challenging to find the best configuration. 
In this paper, we take advantage of the grasp-relevant information $\Omega = \{o, s^{*}, p_{o}\}$ to define a pre-grasp configuration and use a local search method to find the grasp configuration with the highest quality. 

In order to infer the relative pose between the target object and robotic hand, we estimate the pose of the object segment presented in ROI.
Hence, the pre-grasp configuration of the robotic hand is defined as follows:
\begin{enumerate}
\item[1] The palm center $p_h$ is set to be a point along the normal vector of the 3D grasp position $p_{o}^{\prime}$ on the object surface. There is an offset $d$ between $p_h$ and $p_{o}^{\prime}$.
\item[2] The hand palm is perpendicular to the surface normal $n$ at grasp point. 
\item[3] The number of the fingers is selected according to the grasp type $s^{*}$. 
\end{enumerate}

Due to the existence of uncertainties, the defined pre-grasp configuration may fail to grasp objects. 
Hence, a local search is used to find the grasp configuration with the highest quality. 
During searching, we use a local transformation and rotation to sample a set of candidate grasps. 
Hence, the search space is a $4$ dimensional space, $S = \{d, \alpha, \beta, \gamma \}$, where $d$ is the offset between the hand palm center $p_h$ and the 3D grasp position $p_{o}^{\prime}$. 
$\{\alpha, \beta, \gamma\}$ denote the rotate angles in the $X$, $Y$ and $Z$ axes of the hand coordinate respectively. 
The search process is implemented in a simulator and all the candidates are evaluated. 
During executing candidate grasps, all the fingers move to contact with the object surface and stop until the contact force is over a threshold. 
A grasp is considered to be feasible if the robotic hand can grasp and lift the object to a certain height $h$ from the tabletop. 
Multiple feasible grasps are found and evaluated with the quality measurement introduced by \citet{liu2004quality}, which measures the normal component of contact forces. 
Finally, the grasp configuration with the highest quality is chosen. 
Algorithm \ref{algo:grasp} shows the process of the grasp planning procedure. 

\begin{algorithm}
\caption{: Attention based visual analysis for grasp planning}
\label{algo:grasp}
\begin{algorithmic}[1]
\BState \textbf{Requires}: a computational saliency model, a grasp type detection model
\BState Acquire an RGB image $I$ of the table scene.
\BState Visual analysis framework returns the grasp-relevant information $\Omega = \{O, s^{*}, p_{o}\}$.
\BState Using the information $\Omega $ to initialize the pre-grasp configuration of the hand.
\BState Using a local search method to find a list of feasible candidate grasps.
\BState Rank all the feasible candidate grasps to find the best one and execute it in the physical simulator. 
\end{algorithmic}
\end{algorithm}


\section{Experimental Results}
\label{sec:result}
\subsection{Dataset and implementation}
\label{subsec:dataset_and_implementation}
\begin{figure}[H]
\centering
\includegraphics[width=0.8\textwidth]{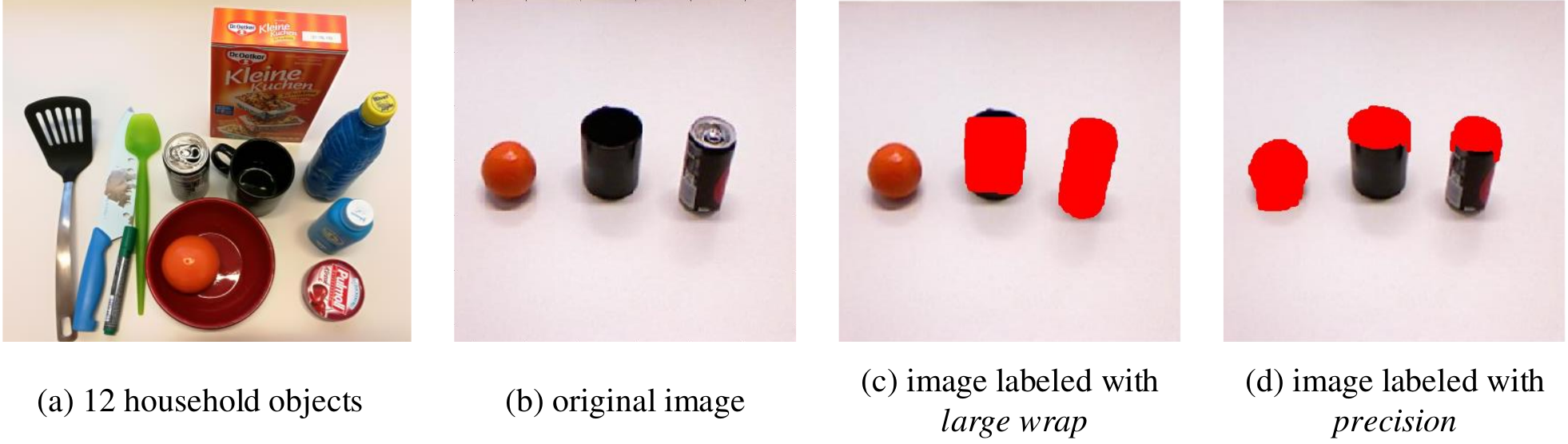}
\caption{Illustration of GTD dataset. (a) 12 household objects contained in the GTD.
(b) The original image. 
(c) A labeled image with \textit{large wrap}.
(d) A labeled image with \textit{precision}.
Pixels that belong to a grasp type are marked with color and others are background}
\label{fig:GTD}
\end{figure}

Existing datasets, such as the Yale human grasping datasets \citep{bullock2015yale} and the UT grasp dataset \citep{cai2015scalable}, are used for the analysis of human grasping behavior, a similar dataset for robotic hands is not available. 
Hence, we introduce a new grasp type detection (GTD) dataset that contains RGB-D \footnote{We use only RGB data in this paper, and plan to exploit the depth data in future work.} images and ground-truth grasp type labels.
There are $11000$ annotated images with resolution $640\times480$.
In this dataset, 6 commonly used grasp types were considered and $12$ household objects with various shape attributes were chosen, as shown in Figure \ref{fig:GTD}.a. 
A MATLAB GUI is designed to manually annotate grasp types on collected data. 
Object parts in images were labeled with different grasp types which enables multi-label detection, as shown in Figure \ref{fig:GTD}(b-c). 
The GTD dataset was split randomly into a training set ($90 \%$) and a testing set ($10\%$). 
The training parameters of the grasp type detection model are set as follows: the initial learning rate was 0.00001 and a step delay policy is used to lower the learning rate as the training progresses.  
Stochastic gradient descent (SGD) method with a momentum rate of 0.9 is used. 

\subsection{Evaluation of grasp type detection}
\label{subsec:eval_grasp}
We first evaluated the accuracy of the grasp type detection on the proposed GTD dataset. 
For comparison, another network based on the Segnet architecture introduced in \citep{badrinarayanan2017segnet} is trained and evaluated. 
Segnet, which is widely used for image segmentation, has an encoder-decoder architecture. 
For pixel-level multi-label detection, we modified the output layer of the Segnet network as introduced in subsection \ref{subsec:grasp_type_detection}. 
The same training and testing procedures are used for both networks described in \ref{subsec:dataset_and_implementation}.
Table \ref{table:accuracy} shows the Intersection-over-union (IoU) of the two networks. 
Our approach achieves a higher average detection accuracy and outperforms the segnet-based network by $10\%$.
\begin{table}[ht]
\centering
\caption{Performance over GTD dataset (IoU).}
\begin{tabular}{c|c|c|c|c|c|c|c}
\toprule
       & L-wrap & S-wrap & Power & Pinch & Precision & Tripod & Average\\
\hline
 Ours  & \textbf{0.63} & \textbf{0.58}  & \textbf{0.71}  & 0.56 & \textbf{0.61}  & \textbf{0.52}  & \textbf{0.60}  \\ 
\hline
 Segnet-based & 0.51 & 0.56 & 0.41 & \textbf{0.61}  & 0.46 & 0.48 & 0.50 \\ 
\bottomrule
\end{tabular}
\label{table:accuracy}
\end{table}

\begin{wrapfigure}{r}{0.5\textwidth}
\centering
\includegraphics[width=0.48\textwidth]{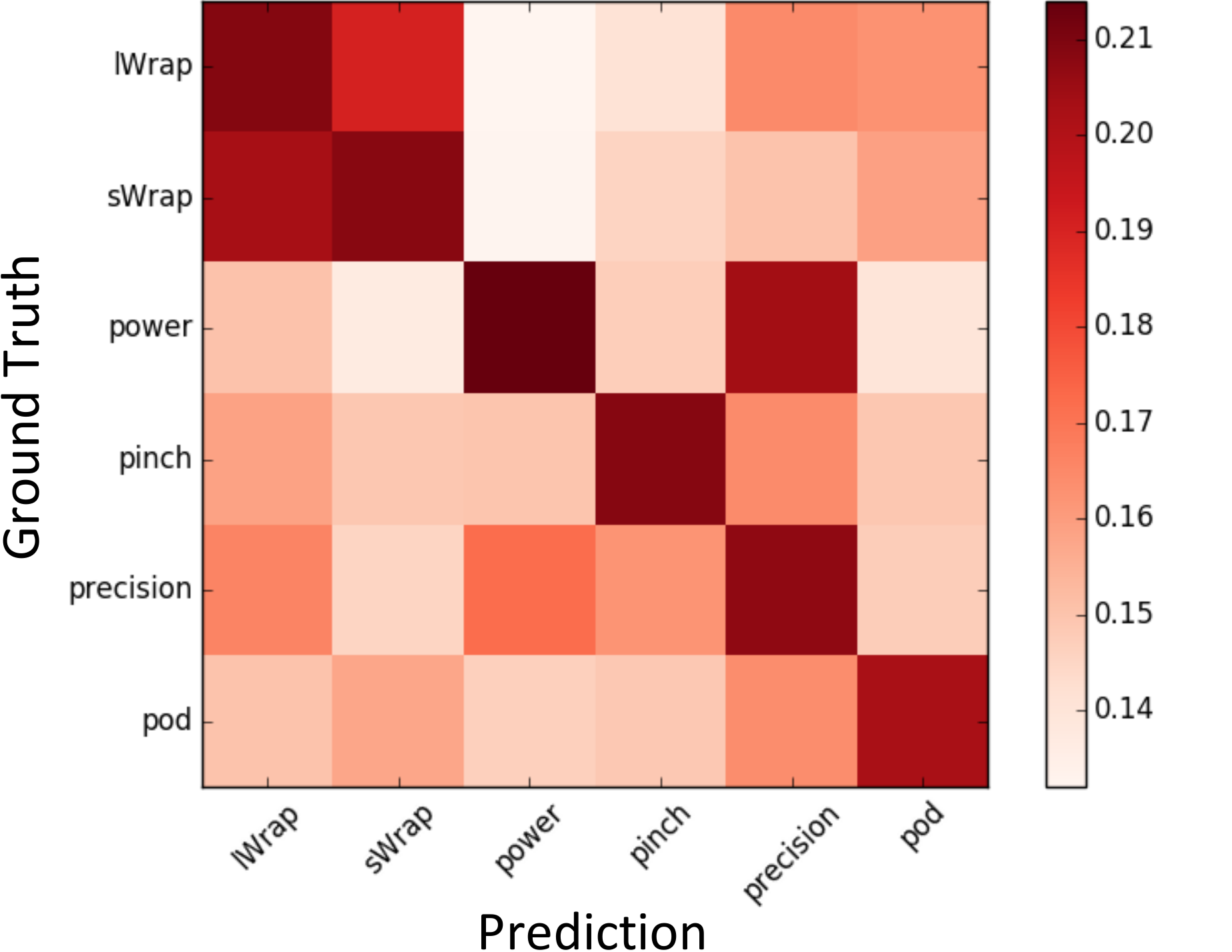}
\caption{The confusion matrix of the six grasp types.}
\label{fig:confusion_matrix}
\end{wrapfigure}

A confusion matrix (Figure \ref{fig:confusion_matrix}) is used to evaluate the overall quality of detected the grasp type.
Since the network predicts 6 labels corresponding to 6 grasp types for each pixel, each row of the matrix shows the predicted probabilities of each grasp type for one ground truth label. 
It shows that the proposed method is able to predict correct grasp types with highest probability since the diagonal elements have the highest values.  
It is worth mentioning that  several off-diagonal elements also have rather high values. 
For example, the prediction results for \textit{Power} type also show a high probability for \textit{Precision}, which means those two grasp types are easily mislabeled by the proposed method.
The reason is that those two types have a high correlation and share many similar characters.
Hence, the confusion matrix gives insights to discover the relationship between grasp types. 

\begin{figure}
\centering
\includegraphics[width=0.75\textwidth]{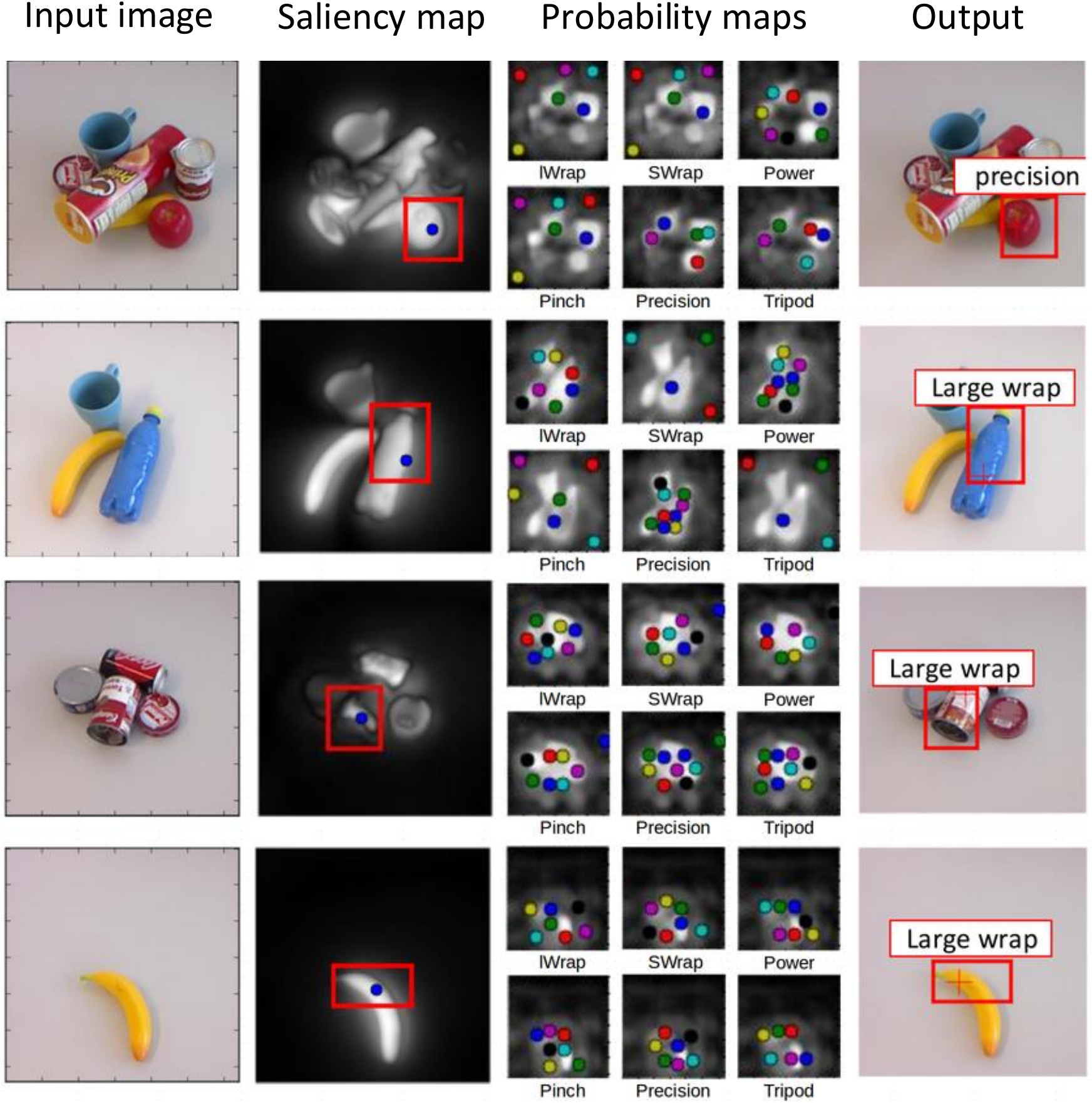}
\caption{Example of the visual analysis on various objects. 
First column is the input RGB image. 
Second column is the pixel-level saliency map, in which the red rectangle denotes the selected ROI. 
Third column is six pixel-level probability maps. 
The color dots in the probability maps denote the cluster centers which are considered as candidate grasp points. 
Last column is the output.}
\label{fig:examples_visual}
\end{figure}

We introduced random objects from the YCB object set \citep{calli2015benchmarking} for testing whether our system can handle unknown objects.
Figure \ref{fig:examples_visual} shows the process of the visual analysis. 
Given an input RGB image, the ROI denoted by a rectangle in the saliency map is first selected by the attention model. 
Meanwhile, six pixel-level probability maps are obtained from the grasp type detection model. 
The possible grasp points denoted by the color dot in each probability map are obtained by clustering. 
Finally, the grasp type with highest probability in the ROI is selected.  
As it is shown in Figure \ref{fig:examples_visual}, our system is also able to produce grasp type and point results on unknown objects. 
For each frame, ROI localization takes $1.8$ seconds, grasp type detection takes $24.18$ seconds and the complete process takes $29.22$ seconds on average.
The proposed framework is implemented in python, and runs on a 2.50GHz Intel i5 CPU.  

\subsection{Grasp planning in simulator}
\label{sebsec:simulation}
The proposed visual analysis framework was further evaluated in object grasping tasks. 
We proposed a grasping simulator based on the V-REP \footnote{http://www.coppeliarobotics.com/}, which is a physical simulator that supports rapid verification, to conduct this experiment. 
The grasping experiments were performed on a shadow hand \footnote{/www.shadowrobot.com/products/dexterous-hand/}, a five-fingered robotic hand which is an approximation of a human hand. 
During simulations, the hand configuration and the contact force between the shadow hand and objects were simulated in real-time, which were used for measuring the qualities of candidate grasps. 

\begin{figure}[H]
\centering
\includegraphics[width=0.8\textwidth]{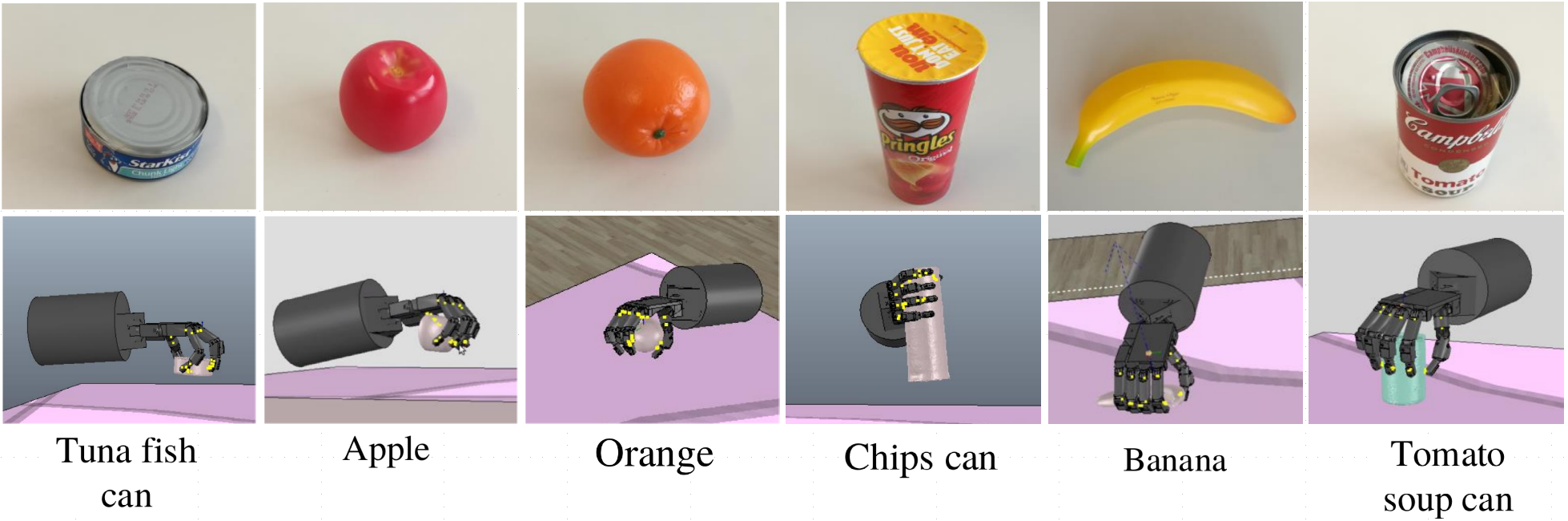}
\caption{Examples of object grasping by the shadow hand in the simulator}
\label{fig:simulation}
\end{figure}

In order to evaluate the performance of the visual analysis framework for the grasp planning, we compared the proposed planning method with the method proposed by \citet{veres2017integrated}. 
Veres et al. used a method which randomly samples a set of candidate grasps based on the normal of the object surface and then ranked all the candidates to find the best one.  
Since there is no grasp type provided in this method, we use the commonly used \textit{power} type for the shadow hand to grasp objects.  
In this comparison experiment, $6$ objects were selected, as shown in Figure \ref{fig:simulation}.  
$10$ trials are tested for each object.  
For each trial, an object is placed on the tabletop and a Kinect sensor captures the RGB image of the table scene.
Then, the grasp configuration of the shadow hand is planned in the simulator.  
The maximum number of search attempts for both methods is limited to 40.   
For each object, the success rate of object grasping and the average number of search attempts needed for finding a feasible grasp are shown in Table \ref{table:simulation}.  

\begin{table}[ht]
\caption{Performance of the proposed grasp planning.}
\centering
\begin{tabular}{c|c|c|c|c}
 \toprule
  & \multicolumn{2}{| c |}{Ours} & \multicolumn{2}{| c }{\citet{veres2017integrated}}\\
\hline
object                & success rate & search attempt & success rate & search attempt \\
\hline
tomato soup can & 8/10 & 2.5 & 8/10 & 20  \\ 
\hline
tuna fish can   & 9/10 & 8.7 & 5/10 & 23.6 \\ 
\hline
banana          & 9/10 & 2.1 & 5/10 & 21.6  \\ 
\hline
apple           & 9/10 & 2.5 & 8/10 & 27.5  \\ 
\hline
orange          & 8/10 & 2.8 & 7/10 & 19.4  \\ 
\hline
chips can       & 10/10 & 2.7 & 10/10 & 11.4 \\ 
\hline
Average         & $88.3\%$ & 3.5 & $71.6\%$ & 20.5 \\ 
\bottomrule
\end{tabular}
\label{table:simulation}
\end{table}

It can be seen that the proposed method obtained a higher success rate of grasping than the random search method. 
Moreover, the number of search attempts by the proposed planning method is only $17.0\%$ compared to that of the random search method. 
It shows that the grasp-relevant information generated helps to reduce the search time needed for the grasp planning and to more accurately find the feasible grasp configuration in the search space.  
It is worth mentioning that the random search method with a \textit{power} type easily fails at grasping some small objects, such as the banana and the tuna fish can. 
This limitation does not occur in the proposed planning method since a feasible grasp type is predicted before grasping. 
Hence, for multi-finger robotic hands, objects with different shape attributes should be handled with different grasp types. 

We also noticed that there are several failures of object grasps using the proposed planning method. 
The main reason for the failures is because the predicted grasp point on the object surface is too close to the tabletop.
Since the environmental constraints are not considered in this work, the shadow hand will collide with the table and fail to grasp the object.
In future, it will be beneficial to also consider environment and task constraints.

\begin{figure}[H]
\centering
\includegraphics[width=0.7\textwidth]{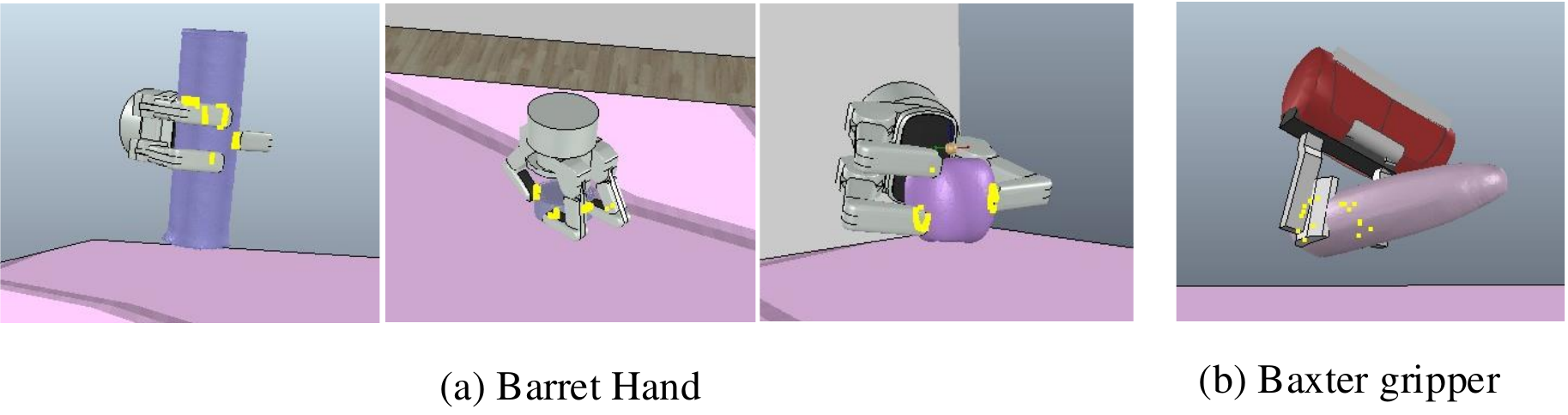}
\caption{Examples of object grasping by the Barrett hand and the Baxter gripper}
\label{fig:other_hand}
\end{figure}

Another state-of-the-art method was proposed by \citet{ciocarlie2009hand}, in which a hand posture subspace is defined for dexterous robotic hands, and the simulated annealing algorithm was used to find a solution in this subspace. 
In this work, the grasp planner only results in a \textit{power} type, which means their grasp planner may fail to grasp small objects. 
Another limitation of their grasp planner is that it needs long search time until finding a feasible solution, with over 70,000 attempts for each plan, and an average running time of 158 seconds \citep{ciocarlie2009hand}. 
Compared with their work, our method requires fewer search attempts. 
We also tested our framework with a 3-fingered Barrett hand \footnote{https://www.barrett.com/about-barrethand/} and a 2-fingered Baxter gripper \footnote{https://www.rethinkrobotics.com/baxter/}, Figure \ref{fig:other_hand} shows some results.
On average, Barrett hand has $90\%$ success rate with 4 search attempts while Baxter gripper has $100\%$ success rate with $1.4$ search attempts. 



\section{Conclusion}
\label{sec:conclusion}
This paper proposes an attention based visual analysis framework, which computes grasp-relevant information directly from visual data for multi-fingered robotic grasping. 
By using the visual framework, a ROI is firstly localized by a computational attention model. 
The grasp type and point on object segment presented in the ROI is then computed using a grasp type detection model, which is used as prior information to guide grasp planning.
We demonstrated that the proposed method is able to give a good prediction of grasp type and point, even in cluttered environments. 
Furthermore, the performance of the proposed visual analysis framework has been evaluated in object grasping tasks. 
Compared to previous methods without prior, the information generated from the visual analysis can significantly speed up the grasp planning. 
Moreover, because using a feasible grasp type, the success rate of the grasping is also improved. 
Results show that the proposed framework helps the robotic systems to know how and where to grasp objects according to attributes of sub-regions of objects.
Since our method does not rely on object detection, it can also handle unseen objects. 

For future work, several aspects are considered:
first, the current framework is goal-driven, and it only learns how to grasp an object, so it will be interesting to extend it into a task-driven framework, e.g. grasping in human-robot handover task;
second, currently the choice of grasp type and point only depends on the attributes of sub-regions of objects. Since grasp planing is also affected by environment and task constraints, those constraints will be taken into consideration; 
finally, we plan to evaluate the proposed framework on real world hardware.



    


\clearpage
\acknowledgments{If a paper is accepted, the final camera-ready version will (and probably should) include acknowledgments. All acknowledgments go at the end of the paper, including thanks to reviewers who gave useful comments, to colleagues who contributed to the ideas, and to funding agencies and corporate sponsors that provided financial support.}


\bibliography{example}   

\end{document}